\documentclass[letterpaper, 10pt, conference]{ieeeconf}  

\IEEEoverridecommandlockouts
\usepackage{graphics}
\usepackage{multirow}
\usepackage{hyperref}
\usepackage{booktabs}
\usepackage{graphicx}
\usepackage{siunitx}
\usepackage{svg}
\usepackage{epsfig}
\usepackage{times}
\usepackage{amsmath}
\usepackage{amssymb}
\usepackage{color}
\usepackage{nicematrix}
\usepackage{url}
\usepackage{algorithm}
\usepackage{algpseudocode}
\usepackage{cite}
\usepackage{bm,bbm}
\usepackage[table]{xcolor}
\usepackage{caption}
\usepackage[labelformat=simple]{subcaption}

\newtheorem{remark}{Remark}

\DeclareMathOperator{\tr}{tr}

\DeclareMathOperator{\SO}{SO}

\newcommand{\R}{\mathbb{R}}

\graphicspath{{./Images/}}
\def\BibTeb{{\rm B\kern-.05em{\sc i\kern-.025em b}\kern-.08em
    T\kern-.1667em\lower.7ex\hbox{E}\kern-.125emb}}

\begin{document}

\title{
Spatiotemporal Calibration of Doppler Velocity Logs for Underwater Robots}

\author{Hongxu Zhao$^{1}$, Guangyang Zeng$^{2}$, Yunling Shao$^{1}$, Tengfei Zhang$^{1}$,Yuzhe Li$^{3}$, and Junfeng Wu$^{1}$
\thanks{$^{1}$The School of Data Science, Chinese University of Hong Kong, Shenzhen, Shenzhen, P. R. China. {\tt\small hxzsimon99@gmail.com, yunlingshao@link.cuhk.edu.cn, zhangtengfei@cuhk.edu.cn,junfengwu@cuhk.edu.cn.} $^{2}$The School of Intelligence Science and Engineering, Harbin Institute of Technology, Shenzhen, Shenzhen 518055, P. R. China.
{\tt\small zengguangyang@hit.edu.cn.}$^{3}$The State Key Laboratory of Synthetical Automation for Process Industries, Northeastern University, Shenyang, P. R. China. {\tt\small yuzheli@mail.neu.edu.cn.}}
}

\maketitle

\begin{abstract}
Accurate fusion of Doppler Velocity Logs (DVLs) with other sensors requires both spatial extrinsics and temporal alignment.
However, existing underwater DVL calibration methods often assume known lever arms and synchronized clocks, and rarely
estimate full calibration parameters jointly.
We propose Unified Iterative Calibration (UIC), a spatiotemporal calibration framework for a DVL and a generic co-sensor
that can provide a continuous-time motion estimate.
We model the base motion with a continuous-time Gaussian process, enabling velocity queries at time-shifted DVL timestamps.
UIC computes a sequential initialization: (i) it estimates the clock offset and DVL scale using low-angular-rate segments via
velocity-norm matching, and (ii) estimates the DVL-to-base rotation and lever arm through a covariance-corrected closed-form
stacked regression with \(\SO(3)\) projection. A nonlinear refinement further reduces residuals.
Simulations with DVL outliers and pool experiments demonstrate robustness and high accuracy downstream odometry estimation.
\end{abstract}

\section{INTRODUCTION}\label{sec:intro}

Doppler Velocity Logs (DVLs) are key sensors for underwater navigation, which can provide body-frame velocity measurements\cite{10342024,10342197,10611064,9721066,webster2010preliminary}.
For multi-sensor fusion (e.g., DVL+camera/SINS), accurate calibration requires both the spatial extrinsics and temporal alignment,
since even a small constant clock offset can cause systematic velocity mismatch under dynamic motion.

Underwater DVL calibration has been studied mainly for DVL--camera systems and for standalone DVLs with navigation-grade
reference velocity sources.
Westman \textit{et al.}~\cite{westman2018underwater} and Xu \textit{et al.}~\cite{xu2021underwater} formulated DVL--camera calibration
as trajectory alignment, but typically rely on known DVL--IMU extrinsics and do not estimate a lever arm and a constant
time offset jointly.
For IMU-free DVLs, works such as \cite{xu2022novel,9069963,9911179,9165098} estimate rotation and scale using SINS/GPS-type velocity references,
which limits applicability when the co-sensor does not directly provide body-frame velocities.
Temporal calibration for heterogeneous sensors has been widely studied on land by fitting continuous-time trajectories and querying
measurements at shifted timestamps (e.g., B-splines/GPs) \cite{6696514,qin2018online,9387269,6631398,7445934}, but has been less explored for DVLs.

\begin{figure}[t]
  \centering
  \includegraphics[width=1\linewidth,height=1\linewidth]{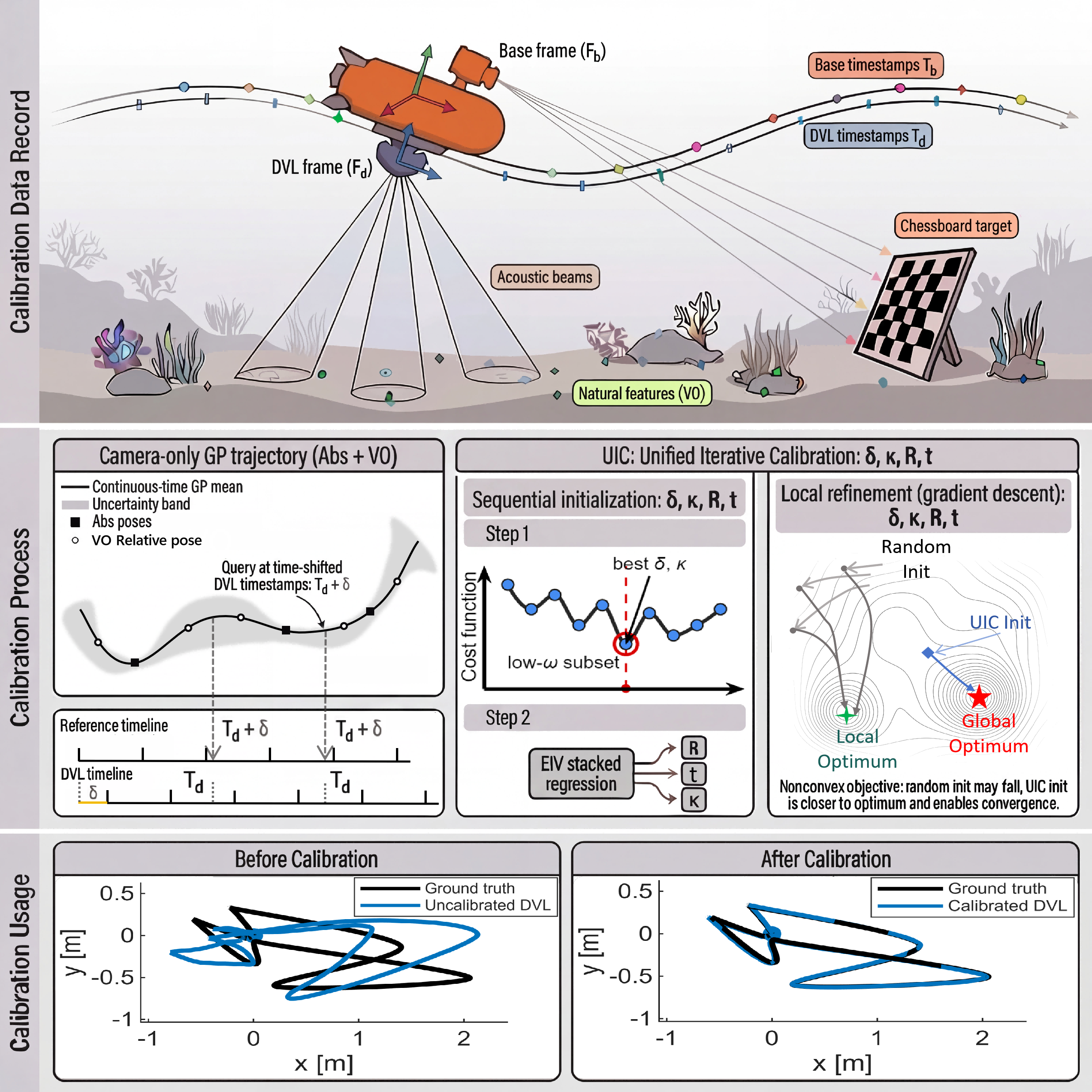}
  \caption{UIC overview: (1) infer a continuous-time base motion with a GP prior, (2) initialize clock offset/scale using low-rotation segments,
  (3) Initialize extrinsics via closed-form stacked regression, and (4) optionally refine \(\Theta\) by local nonlinear least squares.}
  \label{fig:uic_overview}
\end{figure}

We address DVL spatiotemporal calibration by estimating
\(\Theta=\{{}^{b}\delta_{T,d},\,\kappa_d,\,{}^{d}\!R_b,\,{}^{d}\!t_b\}\),
where \({}^{b}\delta_{T,d}\) is a constant clock offset (base time minus DVL time),
\(\kappa_d\) is the DVL velocity scale factor, and \(({}^{d}\!R_b,{}^{d}\!t_b)\) are the rigid extrinsics.
Our formulation does not require the co-sensor to output velocities; it only needs to support a continuous-time motion estimate
from which base-frame linear and angular velocities can be queried.

Our main contributions are:
\begin{itemize}
    \item \textbf{An EIV spatiotemporal calibration model with large-sample characterization under idealized assumptions.}
    We cast DVL spatiotemporal calibration as an error-in-variables (EIV) nonlinear least-squares problem under a continuous-time GP motion prior, jointly estimating
    \(\Theta=\{{}^{b}\delta_{T,d},\,\kappa_d,\,{}^{d}\!R_b,\,{}^{d}\!t_b\}\) via GP-based time-shifted velocity querying.
    Under idealized modeling assumptions and sufficiently exciting motion, we characterize the large-sample behavior of the proposed sequential estimators, establishing bias elimination for all calibration variables.

    \item \textbf{A decoupled calibration pipeline with principled sequential initialization {and closed-form extrinsics + numerical refinement}.}
    We explicitly decouple the clock offset \({}^{b}\delta_{T,d}\) from the spatial extrinsics to reduce mutual interference in joint optimization:
    leveraging (i) the rotation-invariance of velocity norms and (ii) near-zero angular-rate segments (\(\omega_b\approx 0\)) to suppress the lever-arm term, we initialize \(({}^{b}\delta_{T,d},\kappa_d)\) with minimal dependence on \(({}^{d}\!R_b,{}^{d}\!t_b)\).
    Fixing the time offset, we then derive a bias-corrected \emph{closed-form} estimator for the spatial extrinsics (with proper \(\SO(3)\) projection), and use it as an initialization for optional finite-difference numerical-gradient descent refinement.

    \item \textbf{FIM-based excitation analysis and trajectory design guidelines.}
    We analyze the Fisher information matrix (FIM) of the proposed spatiotemporal calibration model to identify the motion excitations required to calibrate all parameters, and translate the analysis into practical trajectory requirements (e.g., combining low-rotation segments for temporal initialization with sufficiently rotational excitation for lever-arm estimation). 
\end{itemize}

{\bf Notation:}~Throughout this paper, we adhere to the following notational conventions: the super-scripted $(\cdot)^o$ represents the true value of a variable $(\cdot)$, $\mathring{(\cdot)}$ a measurement of $(\cdot)$, and $\hat{(\cdot)}$ an estimate to $(\cdot)$. For a vector $a \in \mathbb{R}^3$, let $a^\wedge \in \mathbb{R}^{3\times 3}$ be the skew-symmetric matrix that satisfies $a \times b = a^\wedge b$ for all $b \in \mathbb{R}^3$, where $\times$ is the vector cross product.

\section{Problem formulation}\label{sec_problem_formula}

\subsection{DVL Measurement Model}\label{sec_DVL_model}

When a DVL is mounted on an underwater robot, it emits acoustic beams and measures the Doppler shifts resulting from the robot's movement relative to its ambient environment, which enables the DVL to calculate a three-dimensional velocity under the assumption of a constant and known acoustic speed.

Mathematically, DVL measurements are resolved relative to its local body-fixed frame $\{\mathcal F_d\}$. To analyze the motion of the DVL with respect to a base frame $\{\mathcal F_b\}$ (usually attached to the other sensor), let $p_{o_d,b}$ denote the coordinates of the origin $o_d$ of $\{\mathcal F_d\}$ expressed in frame $\{\mathcal F_b\}$. Essentially, the DVL measures the linear velocity of $o_d$ expressed in its own frame $\{\mathcal F_d\}$. Provided that the base frame and the DVL frame are rigidly connected, when the base moves with linear velocity \(v_{b}\) and angular velocity \(\omega_{b}\) (both expressed in \(\{\mathcal F_b\}\)), the relationship between the DVL velocity \(v_d\) and \((v_b,\omega_b)\) is
\begin{equation}\label{eqn:dvl_model_noise_free}
{v}_{d} = {}^{d}R_{b}\bigl( {v}_{b} + \omega_{b} \times {}^{d}t_{b}\bigr),
\end{equation}
where \({}^{d}R_{b}\in\SO(3)\) is the rotation mapping coordinates from \(\{\mathcal{F}_b\}\) to \(\{\mathcal{F}_d\}\), and \({}^{d}t_{b}\in\mathbb R^3\), which equals to \(p_{o_d,b}\), is the lever arm from the base-frame origin to the DVL-frame origin {expressed in} $\{\mathcal F_b\}$.

To account for sound-speed mismatch (e.g., due to temperature and salinity) and other multiplicative scale biases, we include a positive scaling factor $\kappa_d$. We also consider an additive measurement noise term. Hence, a real DVL measurement $\mathring{v}_{d}$ can be modeled as
\begin{equation}\label{eqn:DVL_measurement_model}
\mathring{v}_{d}
= \kappa_{d}\,{}^{d}\!R_{b}\bigl(\,v_{b} + \omega_{b} \times {}^{d}\!t_{b}\bigr) + n_{v},
\end{equation}
where $n_{v} \sim \mathcal{N}(0,Q_d)$ denotes the measurement noise whose covariance can be obtained from the DVL sensor.

\subsection{Unified Spatiotemporal Calibration Problem under GP Motion Prior}\label{sec_unified_problem}

The robot motion relative to a fixed world frame is represented by a continuous-time state $x(t)$ expressed in the base frame $\{\mathcal F_b\}$. We model $x(t)$ as a Gaussian process (GP),
\begin{equation}\label{eq:gp_prior}
x(t)\sim \mathcal{GP}(\mu(t),K(t,t')),
\end{equation}
where \(K(t,t')\triangleq \operatorname{cov}\!\bigl(x(t),x(t')\bigr)\) is the (matrix-valued) GP covariance kernel (with chosen hyperparameters)~\cite{williams2006gaussian}.The reference sensor suite attached to the base provides measurements $\mathring{\mathcal M}_b=\{\mathring m_{b,j}\}_{j=1}^M$ time-stamped at ${}^bT=\{{}^bT_j\}_{j=1}^M$, with measurement model
\begin{equation}\label{Measurements_onB}
\mathring m_{b,j}=h\!\big(x({}^bT_j)\big)+n_{b,j},~ n_{b,j}\sim\mathcal{N}(0,R_{b,j}),~ j=1,\dots,M.
\end{equation}
Define the stacked vector $\mathring M_b \triangleq [\mathring m_{b,1}^\top~\cdots~\mathring m_{b,M}^\top]^\top$ and let $H$ denote the Jacobian of $h(\cdot)$ evaluated at the linearization point used by GP regression. Let $X_b \triangleq [x({}^bT_1)^\top~\cdots~x({}^bT_M)^\top]^\top$ be the stacked state at the base timestamps. Using $\mathring M_b$ and the GP prior, we compute a reference trajectory estimate (and its uncertainty) via GP regression, denoted by $\hat X_b$ (and $\Sigma_{X_b}$); the closed-form expressions are provided in Section~\ref{section:Initialization_X}.

The DVL provides velocity measurements $\mathring{\mathcal V}_d=\{\mathring v_{d,i}\}_{i=1}^N$ time-stamped by its local clock at ${}^dT=\{{}^dT_i\}_{i=1}^N$. A constant clock offset between the base and DVL clocks is modeled by ${}^{b}\delta_{T,d}$, such that the physical time of a DVL sample recorded at ${}^dT_i$ corresponds to the base time ${}^dT_i+{}^{b}\delta_{T,d}$. The calibration variables of interest are collected in
\[
\Theta \triangleq \bigl\{{}^{b}\delta_{T,d},\,{}^{d}\!R_{b},\,{}^{d}\!t_{b},\,\kappa_{d}\bigr\},
\]
where ${}^{d}\!R_{b}\in\SO(3)$ and ${}^{d}\!t_b\in\mathbb{R}^3$ are the extrinsic parameters, and $\kappa_d>0$ is the DVL velocity scale factor. For a candidate ${}^{b}\delta_{T,d}$, define the aligned DVL timestamps in the base clock by ${}^bT_{d,i}({}^{b}\delta_{T,d})\triangleq {}^dT_i+{}^{b}\delta_{T,d}$, and let $X_d({}^{b}\delta_{T,d}) \triangleq
[x({}^bT_{d,1}({}^{b}\delta_{T,d}))^\top~\cdots~x({}^bT_{d,N}({}^{b}\delta_{T,d}))^\top]^\top$
denote the queried states at these aligned instants. Given $\hat X_b$, the queried mean $\hat X_d({}^{b}\delta_{T,d})$ is obtained by GP interpolation (as an explicit function of ${}^{b}\delta_{T,d}$), and the corresponding queried covariance $\Sigma_{X_d}({}^{b}\delta_{T,d})$ can be computed as a byproduct of GP regression and interpolation following ~\eqref{interpolation} and~\eqref{uncertainty}.

At aligned time ${}^bT_{d,i}({}^{b}\delta_{T,d})$, we expect the DVL measurement as $\kappa_{d}\,{}^{d}\!R_{b}\bigl(\hat v_{b,i}({}^{b}\delta_{T,d}) + \hat\omega_{b,i}({}^{b}\delta_{T,d}) \times {}^{d}\!t_{b}\bigr)$,
where $\hat v_{b,i}({}^{b}\delta_{T,d})$ and $\hat\omega_{b,i}({}^{b}\delta_{T,d})$ are the linear and angular velocity components extracted from the queried state $\hat X_{d,i}({}^{b}\delta_{T,d})$. We then define the DVL least-squares cost
\begin{equation}\label{eq:JdM_def_ls}
\begin{aligned}
J_{d,\mathrm{M}}\bigl(\hat X_d({}^{b}\delta_{T,d}),{}^{d}\!R_{b},{}^{d}\!t_{b},\kappa_{d}\bigr)
\triangleq&\\
\sum_{i=1}^N
\Bigl\|
\mathring{v}_{d,i}
-
\kappa_{d}\,{}^{d}\!R_{b}\bigl(\hat v_{b,i}({}^{b}\delta_{T,d}) + &\hat\omega_{b,i}({}^{b}\delta_{T,d}) \times {}^{d}\!t_{b}\bigr)
\Bigr\|_{Q_d}^{2}.
\end{aligned}
\end{equation}
We pose spatiotemporal calibration as the following constrained nonlinear least-squares problem:
\begin{align}
\operatorname*{min}_{\substack{
{}^{b}\delta_{T,d},\,{}^{d}\!R_{b},\,{}^{d}\!t_{b},\,\kappa_{d}
}}
\;\;&
J_{d,\mathrm{M}}\bigl(\hat X_d({}^{b}\delta_{T,d}),{}^{d}\!R_{b},{}^{d}\!t_{b},\kappa_{d}\bigr)
\notag\\
\text{s.t.}\quad
& {}^{b}\delta_{T,d} \in \mathbb R, {}^{d}\!R_b \in \SO(3), {}^{d}\!t_b \in \R^3, \kappa_{d}>0.
\label{problem:MAP_calibr}
\end{align}

Problem~\eqref{problem:MAP_calibr} is an error-in-variables (EIV) problem~\cite{wald1940fitting,mu2014recursive,reiersol1950identifiability}, because the queried trajectory $\hat X_d({}^{b}\delta_{T,d})$ is itself estimated from noisy reference measurements but is independent of the DVL noise. The problem is also highly nonconvex owing to the coupled spatiotemporal parameters and time-shifted GP queries. We therefore develop a sequential estimator of $\Theta$ to initialize the subsequent local nonlinear least-squares refinement.

\section{METHODOLOGY}\label{Methdodlogy}
UIC consists of sequential initialization followed by local iterative refinement. The initializer estimates all components of $\Theta$ without manual or near-true guesses. The reference trajectory $x(t)$ is inferred from the base measurements under the GP prior and is not jointly optimized with $\Theta$, because a flexible trajectory could absorb spatiotemporal miscalibration and confound parameter estimation. Under the idealized EIV assumptions, the initializer is designed to enter a neighborhood of the true parameters, after which refinement further decreases the residual.

\subsection{Sequential Initialization Strategy}

Our initialization procedure separates reference motion estimation from the calibration parameters ${}^{d}\!R_{b}$, ${}^{d}\!t_{b}$, $\kappa_{d}$, and ${}^{b}\delta_{T,d}$. Concretely, UIC first computes a reference trajectory estimate $\hat X_b$ and its uncertainty $\Sigma_{X_b}$ from the base measurements $\mathring M_b$ under the GP motion prior. Then, for any candidate clock offset ${}^{b}\delta_{T,d}$, the time-shifted motion states at the aligned DVL timestamps are obtained by GP interpolation, yielding $\hat X_d({}^{b}\delta_{T,d})$ and $\Sigma_{X_d}({}^{b}\delta_{T,d})$. This stage does not use DVL measurements, so that the reference motion estimate remains independent of the unknown calibration parameters and admits explicit GP-based interpolation. According to~\cite{doi:10.1177/0278364913478672}, the GP regression/interpolation stage can be implemented efficiently and provides both a reference motion estimate and an uncertainty characterization, which are useful for subsequent calibration. The large-sample analysis is derived under idealized modeling assumptions (e.g., correct GP regression/interpolation statistics and sufficiently excited motion), and is intended to justify the design of the initialization estimators rather than to serve as finite-sample performance guarantees.

Directly optimizing Problem~\eqref{problem:MAP_calibr} from random initialization offers no strong convergence guarantees. Poor initial guesses often lead to divergence or convergence to suboptimal local minima. In our experiments, naive initialization frequently resulted in non-convergence. The initialization procedures are presented in the next two subsections.

\subsection{Initialization of Motion State $X$}\label{section:Initialization_X}

By GP regression conditioned on the base measurements \(\mathring M_b\) and the GP prior, we obtain an estimate \(\hat X_b\) of the base states \(X_b\) in closed form. Then, for any candidate clock offset \({}^{b}\delta_{T,d}\), the queried states at the aligned DVL timestamps are computed as \(\hat X_d({}^{b}\delta_{T,d})\) via standard GP interpolation. Let \({}^{b}T_{d}({}^{b}\delta_{T,d})\triangleq\{{}^{b}T_{d,i}({}^{b}\delta_{T,d})\}_{i=1}^{N}\), where
\({}^{b}T_{d,i}({}^{b}\delta_{T,d})\triangleq{}^{d}T_i+{}^{b}\delta_{T,d}\).
By abuse of notation, \(K({T}_1,{T}_2)\) denotes the block (cross-)covariance matrix assembled from the GP kernel \(K(t,t')\).
Define \(K_1\triangleq K({}^{b}T,{}^{b}T)\), \(K_2({}^{b}\delta_{T,d})\triangleq K({}^{b}T,{}^{b}T_{d}({}^{b}\delta_{T,d}))\),
\(K_3({}^{b}\delta_{T,d})\triangleq K({}^{b}T_{d}({}^{b}\delta_{T,d}),{}^{b}T)\), and \(K_4({}^{b}\delta_{T,d})\triangleq K({}^{b}T_{d}({}^{b}\delta_{T,d}),{}^{b}T_{d}({}^{b}\delta_{T,d}))\). Similarly $\mu=\begin{bmatrix}
    \mu_1\\\mu_2({}^{b}\delta_{T,d})
\end{bmatrix}$ which $\mu_1$ denote the GP prior mean at $^bT$ and $\mu_2$ denote the GP prior mean at ${}^{b}T_{d}({}^{b}\delta_{T,d})$.
According to the linear prediction equations~\cite[(50)]{doi:10.1177/0278364913478672}, \(\hat X_b\) is estimated via GP regression as 
\begin{equation}\label{eqn:xb_closed_form}
\hat X_b=\mu_1+K_1H^\top(R+HK_1H^\top)^{-1}(M_b-H\mu_1).
\end{equation}
Then \(\hat X_d\) is computed as a linear combination of \(\hat X_b\), with weights determined explicitly by \({}^bT\), \({}^dT\), and \({}^b\delta_{T,d}\), i.e.,
\begin{equation}\label{interpolation}
\hat X_d({}^b\delta_{T,d})=\mu_2({}^b\delta_{T,d})-K_3({}^b\delta_{T,d})K_1^{-1}(\mu_1-\hat X_b).
\end{equation}
This makes \(\hat X_d({}^{b}\delta_{T,d})\) an explicit function of the time offset through the GP interpolation weights. In practice, the GP regression in~\eqref{eqn:xb_closed_form} and the query in~\eqref{interpolation} can be implemented either in closed form for moderate problem sizes, or via sparse/Markov representations when the chosen kernel admits a state-space form; both filtering/smoothing and factor-graph solvers are applicable. In addition, the posterior uncertainty can be computed as a byproduct of GP regression/interpolation~\cite{williams2006gaussian}, given by
\begin{equation*}
\Sigma_{X_b}=K_1-K_1H^\top(R+HK_1H^\top)^{-1}HK_1,
\end{equation*}
\begin{equation}\label{uncertainty}
\begin{aligned}
\Sigma_{X_d}({}^b\delta_{T,d})=&K_4({}^b\delta_{T,d})-
\\&K_3({}^b\delta_{T,d})K_1^{-1}\!(K_1-\Sigma_{X_b})K_1^{-1}\!K_2({}^b\delta_{T,d}).
\end{aligned}
\end{equation}

\subsection{Initialization of Calibration Variables}\label{section:Initialization_extrincis}
We next initialize the calibration variables using all DVL measurements together with the GP estimate $\hat X_d({}^b\delta_{T,d})$. 
We decouple the calibration of the extrinsic parameters \({}^{d}\!R_b\) and \({}^{d}\!t_b\) from that of the temporal offset and scale parameters \({}^{b}\delta_{T,d}\) and \(\kappa_d\) by leveraging specialized robot motion patterns. As will be shown shortly, the resulting estimators can be computed efficiently---via line search for the temporal/scale parameters, and a closed-form bias-corrected estimator for the spatial extrinsics. 

\subsubsection{Clock offset and scale initialization}\label{section:temporal_calibration}
In principle, we estimate the clock offset \( {}^{b}\delta_{T,d} \) by maximizing the alignment between the DVL measurements and the predicted DVL output based on \( \hat{X}_d({}^{b}\delta_{T,d}) \). Specifically, the predicted DVL velocity is given by $\kappa_d\, {}^{d}\!R_b \left( \hat v_{b,i}({}^{b}\delta_{T,d}) + \hat \omega_{b,i}({}^{b}\delta_{T,d}) \times {}^{d}\!t_b \right)$
according to~\eqref{eqn:DVL_measurement_model}. To reduce the influence of the spatial extrinsics in this stage, we select a subset of measurements \(\mathcal D_{o}\) with small angular velocity, i.e., \(\omega_{b,i}\approx 0\). Under this operating regime, the lever-arm term \(\hat\omega_{b,i}({}^{b}\delta_{T,d}) \times {}^{d}\!t_b\) becomes negligible. Furthermore, taking the Euclidean norm removes the effect of \({}^{d}\!R_b\) due to rotation invariance. Combining these yields, for \(i\in\mathcal D_o\),
\begin{align}\label{eqn:angular_velocity_approx}
\kappa_d^2\|\hat v_{b,i}({}^{b}\delta_{T,d})\|^2
&\approx \|v_{d,i}^{o}\|^2.
\end{align}
Centering the noise terms leads to the following two-dimensional search problem:
\begin{equation}\label{clock_offset_calibration}
\mathop{\rm min}\limits_{{}^{b}\delta_{T,d}\kappa_{d}}
\!\!\!\sum_{i\in\mathcal D_{o}}
\!\!\Bigl(\!
\dfrac{\|\mathring{v}_{d,i}\|^2\!\!-\!\tr(Q_d)}{\kappa_d^2}
\!-\!
\|\hat v_{b,i}({}^{b}\delta_{T,d})\|^2
\!+
\tr\!\bigl(\Sigma_{v_b,i}({}^{b}\delta_{T,d})\!\bigr)
\!\!\Bigr)^2\!\!\!,
\end{equation}
where $\Sigma_{v_b,i}({}^{b}\delta_{T,d})$ denotes the $3\times 3$ marginal covariance of the interpolated linear velocity $\hat v_{b,i}({}^{b}\delta_{T,d})$, extracted from $\Sigma_{X_d}({}^{b}\delta_{T,d})$. The problem can be efficiently solved by performing a grid search over ${}^{b}\delta_{T,d}$ while varying $\kappa_d$ over a small grid around $1$. Let ${}^{b}\hat \delta_{T,d}$ and $\hat \kappa_d$ be any global minimizer. The theorem characterizes  large-sample behavior under the idealized regime \(\omega_{b,i}= 0\) for \(i\in\mathcal D_o\).

\begin{remark}\label{remark1}
Under idealized conditions (e.g., sufficiently small angular rates on the selected subset, correct second-moment models for GP interpolation errors, and identifiable speed excitation), the minimizer of~\eqref{clock_offset_calibration} is a consistent estimator of \(({}^{b}\delta_{T,d},\kappa_d)\) as the number of selected samples increases.
This remark follows standard EIV-estimation and is included to motivate the estimator structure rather than to provide a finite-sample guarantee.
\end{remark}


\subsubsection{Bias-corrected closed-form extrinsics initialization}

Stacking all DVL measurements as $\mathring{V}_d = [\mathring{v}_{d,1}\cdots \mathring{v}_{d,N}]^\top$, we arrive at the stacked version of~\eqref{eqn:DVL_measurement_model}:
\begin{equation}
\mathring{V}_d = \kappa_d V_b^o  {}^d\!R_b^{\top}
- \kappa_d \Omega_b^o ({}^d\!t_b^{\wedge})^{\top} {}^d\!R_b^{\top} + N_{v},\label{Matrix_measurement}
\end{equation}
where $V_b^o = [ v_{b,1}^{o}\cdots v_{b,N}^{o}]^\top$, $\Omega_b^o =[ \omega_{b,1}^{o}\cdots \omega_{b,N}^{o}]^\top$ and $N_{v} = [ n_{v,1}\cdots n_{v,N}]^\top$.
Define
\[
Z_1=\kappa_{d}\,{}^{d}\!R_b^{\top},~~
Z_2=\kappa_{d}\,({}^{d}\!t_b)^{\wedge\top}\,{}^{d}\!R_b^{\top},~~
Z\triangleq
\begin{bmatrix}
Z_1\\
Z_2
\end{bmatrix}\in\mathbb R^{6\times 3},
\]
and define \(H^o\triangleq[V_{b}^o~-\Omega_{b}^o]\in\mathbb R^{N\times 6}\).
Then~\eqref{Matrix_measurement} can be written compactly as
\(\mathring{V}_{d}=H^oZ+N_{v}\).
Omitting the $\SO(3)$ constraint of $^{d}\!R_b$, we estimate $Z$ by solving
\begin{equation*}
\underset{Z}{\operatorname{argmin}}~~\|\mathring{V}_{d}-H^oZ\|_F^2,
\end{equation*}
which admits the closed-form solution
\begin{equation}\label{eqn:Closed_form_Z}
\hat{Z}=\Bigl(\frac{H^{o\top} H^o}{N}\Bigr)^{-1}\Bigl(\frac{H^{o\top} \mathring{V}_{d}}{N}\Bigr).
\end{equation}
We note that \(\kappa_d\) is estimated again in the above problem. This is a deliberate technical choice: the earlier estimation used only a selected subset \(\mathcal{D}_o\) of DVL measurements with \(\omega_{b,i} \approx 0\), which represents a relatively limited volume of data. In contrast, the stacked formulation~\eqref{Matrix_measurement} uses the entire measurement set. This yields a reliable estimate of \(\kappa_d\) in the sense that it exhibits less variance around the true value.

The exact values of \(v_{b,i}^o\) and \(\omega_{b,i}^o\) are not directly measurable, therefore their estimates \(\hat{v}_{b,i}\) and \(\hat{\omega}_{b,i}\) from \(\hat{X}_b\) are substituted into~\eqref{eqn:Closed_form_Z}. Accordingly, \(\hat{Z}\) should be corrected to account for this substitution in the EIV setting. The resulting bias-corrected estimate is
\begin{equation}\label{closed_form_zc}
\hat{Z}^{c} = \Bigl( \frac{\hat H^\top \hat H - G}{N} \Bigr)^{-1}
\Bigl( \frac{\hat H^\top \mathring{V}_{d}}{N} \Bigr),
\end{equation}
where $\hat H=[\hat V_{b}~-\hat \Omega_{b}]$ with $\hat V_{b}=[\hat v_{b,1}( {}^{b}\hat \delta_{T,d})\cdots \hat v_{b,N}({}^{b}\hat \delta_{T,d})]$ and $\hat \Omega_{b}=[\hat \omega_{b,1}( {}^{b}\hat \delta_{T,d})\cdots \hat \omega_{b,N}({}^{b}\hat \delta_{T,d})]$, and $G\in\R^{6\times 6}$ is the covariance-induced correction term. Specifically, letting $\Sigma_{v_b,i}({}^{b}\hat \delta_{T,d})$ and $\Sigma_{\omega_b,i}({}^{b}\hat \delta_{T,d})$ denote the $3\times 3$ marginal covariances of the interpolated linear and angular velocities (extracted from \(\Sigma_{X_d}({}^{b}\hat \delta_{T,d})\) in~\eqref{uncertainty}), we define
\[
G \triangleq \sum_{i=1}^N \mathrm{blkdiag}\!\bigl(\Sigma_{v_b,i}({}^{b}\hat \delta_{T,d}),~\Sigma_{\omega_b,i}({}^{b}\hat \delta_{T,d})\bigr).
\]

We further reconstruct $\hat{\kappa}_d^c$ from $\hat{Z}^{c}$ as
\begin{equation}\label{RecoverRT}
\hat{\kappa}_d^c=\bigl|{\rm det}(\hat{Z}_1^c)\bigr|^{1/3},
\end{equation}
which follows from $\det(^d\!\hat{R}_b^o)=1$.
Performing singular value decomposition (SVD) on $(\hat{Z}_1^{c}/\hat{\kappa}_d^c)^{\top}$, we write
$(\hat{Z}_1^{c}/\hat{\kappa}_d^c)^{\top}=U_R {\Sigma}_R {V}_R^\top$, and set
\begin{equation}\label{SVD}
{}^{d}\!\hat{R}_b^c
=
U_R \, \mathrm{diag}\!\bigl([1,1,\det(U_R V_R^{\top})]\bigr) V_R^\top .
\end{equation}
We then recover an estimate ${}^{d}\!\hat{t}_{b}^{c{\wedge}}$ from $\hat{Z}^{c}_{2}$ as
\begin{equation}\label{closed_form_t}
{}^{d}\!\hat{t}_{b}^{c{\wedge}}
=
\Bigl(
-\hat{Z}^{c}_{2}\,{}^{d}\!\hat{R}_b^c/\hat{\kappa}_d^c
-
\bigl(-\hat{Z}^{c}_{2}\,{}^{d}\!\hat{R}_b^c/\hat{\kappa}_d^c\bigr)^\top
\Bigr)/2.
\end{equation}

The following result characterizes the large-sample behavior of the above estimates under standard EIV conditions.

\begin{remark}\label{remark2}
When the motion is sufficiently exciting so that the normal matrix is nonsingular, and when the GP-interpolated velocity errors satisfy standard regularity conditions, the structured estimator recovered from~\eqref{closed_form_zc}--\eqref{closed_form_t} is consistent under standard EIV arguments.
This remark is primarily included to motivate the bias-correction structure; finite-sample behavior is assessed experimentally.
\end{remark}

\subsection{Iterative Refinement}\label{sec:refine}
Starting from the sequential initialization \(\hat\Theta\), we refine \(\Theta\) by local iterative optimization of Problem~\eqref{problem:MAP_calibr}.
The rotation is updated on \(\SO(3)\) via an exponential-map perturbation, and the DVL scale is re-parameterized to enforce \(\kappa_d>0\):
\begin{align}
{}^{d}\!R_b &\leftarrow \exp(\phi^\wedge)\,{}^{d}\!R_b,\qquad \phi\in\mathbb{R}^3,\\
\kappa_d &= \exp(s_d),\qquad s_d\in\mathbb{R}.
\end{align}
We apply numerical gradient descent to refine the parameters.

\subsection{Trajectory Design}\label{sec:traj_design}
The calibration accuracy of
\(\Theta\)
depends critically on motion excitation, which can be characterized via a Fisher information matrix (FIM)
induced by the DVL velocity residual model in~\eqref{eq:JdM_def_ls}.
Importantly, the predicted body-frame motion \(\hat v_b(\cdot)\) and \(\hat\omega_b(\cdot)\) is obtained from the continuous-time GP
and therefore carries uncertainty.
To reflect this, we use an \emph{effective} residual covariance \(Q_e\), which captures not only the intrinsic DVL measurement noise,
but also the uncertainty induced by the estimated motion signals \(\hat v_b,\hat\omega_b\).

Define the predicted DVL velocity at aligned time
\({}^bT_{d,i}({}^{b}\delta_{T,d})={}^dT_i+{}^{b}\delta_{T,d}\) as
\[
\bar v_{d,i}(\Theta)
=
\kappa_d\,{}^{d}\!R_b
\Bigl(
\hat v_{b,i}({}^{b}\delta_{T,d})
+
\hat\omega_{b,i}({}^{b}\delta_{T,d})\times{}^{d}\!t_b
\Bigr).
\]
Denote \(r_i(\Theta)\triangleq \mathring v_{d,i}-\bar v_{d,i}(\Theta)\) and
\(y_i({}^{b}\delta_{T,d})\triangleq \hat v_{b,i}({}^{b}\delta_{T,d})+\hat\omega_{b,i}({}^{b}\delta_{T,d})\times{}^{d}\!t_b\).
Using the unconstrained scale parameterization \(\kappa_d=\exp(s_d)\) and a left perturbation
\({}^{d}\!R_b\leftarrow \exp(\phi^\wedge)\,{}^{d}\!R_b\),
the first-order residual Jacobians satisfy
\begin{align*}
J_{s,i}
&\triangleq
\frac{\partial r_i}{\partial s_d}
=
-\kappa_d\,{}^{d}\!R_b\,y_i({}^{b}\delta_{T,d}),\\
J_{\phi,i}
&\triangleq
\frac{\partial r_i}{\partial \phi}
=
\kappa_d\bigl({}^{d}\!R_b\,y_i({}^{b}\delta_{T,d})\bigr)^\wedge,\\
J_{t,i}
&\triangleq
\frac{\partial r_i}{\partial({}^{d}\!t_b)}
=
-\kappa_d\,{}^{d}\!R_b\bigl(\hat\omega_{b,i}({}^{b}\delta_{T,d})\bigr)^\wedge,\\
J_{\delta,i}
&\triangleq
\frac{\partial r_i}{\partial {}^{b}\delta_{T,d}}
=
-\kappa_d\,{}^{d}\!R_b
\Bigl(
\dot{\hat v}_{b,i}({}^{b}\delta_{T,d})
+
\dot{\hat\omega}_{b,i}({}^{b}\delta_{T,d})\times{}^{d}\!t_b
\Bigr),
\end{align*}
where \(\dot{(\cdot)}\) denotes a time derivative evaluated at the aligned instant, which is always available analytically under a continuous-time GP model.

The FIM can be written in the stacked form
\[
\mathcal I(\Theta)
=
J^\top Q_e^{-1}J,
~
J\triangleq
\begin{bmatrix}
J_1\\
\vdots\\
J_N
\end{bmatrix},
~
J_i\triangleq [\,J_{\delta,i}\;\;J_{s,i}\;\;J_{\phi,i}\;\;J_{t,i}\,].
\]
Here \(Q_{e}\) is the effective covariance of the residuals, which aggregates the DVL measurement uncertainty and the additional uncertainty
caused by motion estimates.

From the FIM, the lever arm \({}^{d}\!t_b\) is excited only by rotation since \(J_{t,i}\propto(\hat\omega_{b,i}({}^{b}\delta_{T,d}))^\wedge\); thus \(\hat\omega_{b,i}({}^{b}\delta_{T,d})\approx 0\) makes \({}^{d}\!t_b\) unobservable, and near single-axis rotation yields an ill-conditioned \(\mathcal I_{tt}\) with the component of \({}^{d}\!t_b\) along that axis weakly observable. The clock offset \({}^{b}\delta_{T,d}\) is mainly informed by translational/rotational \emph{accelerations} through \(J_{{}^{b}\delta_{T,d},i}\), so nearly constant-speed and constant-turn-rate motions are uninformative for \({}^{b}\delta_{T,d}\), while \(({}^{d}\!R_b,\kappa_d)\) benefit from large, direction-rich \(y_i({}^{b}\delta_{T,d})\). Accordingly, we recommend a \emph{multi-regime} calibration trajectory: (a) low angular rate with deliberately varying speed/direction to excite \({}^{b}\delta_{T,d}\) while suppressing lever-arm coupling, and (b) sufficiently rich multi-axis rotation with nonzero translation to excite \({}^{d}\!t_b\) and keep \(y_i({}^{b}\delta_{T,d})\) informative for \({}^{d}\!R_b\) and \(\kappa_d\).

\begin{algorithm}[t]
\renewcommand{\algorithmicrequire}{\textbf{Input:}}
\renewcommand{\algorithmicensure}{\textbf{Output:}}
\caption{{UIC: Sequential Initialization + Gradient-Descent Refinement}}
\label{DVL_algorithm}
\begin{algorithmic}[1]
\Require time-stamped DVL measurements \(\{(\mathring v_{d,i},T_{d,i})\}_{i=1}^{N}\);
time-stamped base measurements \(\{(\mathring m_{b,j},T_{b,j})\}_{j=1}^{M}\);
GP kernel \(K(t,t')\) in~\eqref{eq:gp_prior};
angular-rate threshold \(\omega_{\rm th}\) for selecting \(\mathcal D_o\)
\Ensure {\({}^{b}\hat\delta_{T,d}\), \(\hat\kappa_{d}\), \({}^{d}\!\hat R_{b}\), \({}^{d}\!\hat t_{b}\)}

\State \textbf{GP reference trajectory:}
compute \(\hat X_{b}(\cdot)\) with covariances from \(\{\mathring m_{b,j}\}\) and \(K\) via~\eqref{eqn:xb_closed_form}--\eqref{uncertainty}.

\State \textbf{Temporal/scale pre-initialization on \(\mathcal D_o\):}
solve~\eqref{clock_offset_calibration} by a grid search over \({}^{b}\delta_{T,d}\) and \(\kappa_d\) around \(1\) only on the low-angular-velocity index set.

\State \textbf{Bias-corrected stacked extrinsics initialization:}
using \({}^{b}\hat\delta_{T,d,\rm init}\), construct \(\hat H\) and \(G\) as in~\eqref{closed_form_zc}, and compute \(\hat Z^c\) using all \(\mathring v_{d,i}\).

\State \textbf{Recover scale and extrinsics:}
compute \(\hat\kappa_{d,\rm init}\), \({}^{d}\!\hat R_{b,\rm init}\), and \({}^{d}\!\hat t_{b,\rm init}\) from \(\hat Z^c\) via~\eqref{RecoverRT},~\eqref{SVD}, and~\eqref{closed_form_t}.

\State {\textbf{Numerical-gradient-descent refinement:}
perform gradient descent on~\eqref{problem:MAP_calibr} with respect to \(\Theta\). Return the final \(\hat\Theta\) as \(\{{}^{b}\hat\delta_{T,d},{}^{d}\!\hat R_{b},{}^{d}\!\hat t_{b},\hat\kappa_{d}\}\).}
\end{algorithmic}
\end{algorithm}

\section{NUMERICAL ANALYSIS}\label{simulation}
This section presents Monte Carlo simulation results that evaluate the proposed UIC under a camera-DVL calibration problem and quantify how spatiotemporal parameters and motion excitation affect DVL calibration and fusion.

\subsection{Simulation setup and evaluation protocol}\label{sec:sim_setup_protocol}

Each trial synthesizes a bounded 3D \(\mathrm{SE}(3)\) trajectory together with three measurement streams:
(i) \emph{absolute poses} (Abs) available only in a ``calibration zone'' and under low-speed/low-turn-rate gating, simulating camera visualizing a calibration board.
(ii) \emph{relative poses} (VO) at a fixed rate simulating visual odometry, and
(iii) \emph{DVL body-frame linear velocities}.
The ground-truth calibration parameters \(\Theta^{o}=\{\delta^{o},\kappa_d^{o},{}^{d}\!R_b^{o},{}^{d}\!t_b^{o}\}\) are fixed across MC runs.

\begin{figure*}[t]
  \centering
  \captionsetup[sub]{justification=centering,singlelinecheck=false}

  \begin{subfigure}[t]{0.245\textwidth}
    \centering
    \includegraphics[width=\linewidth]{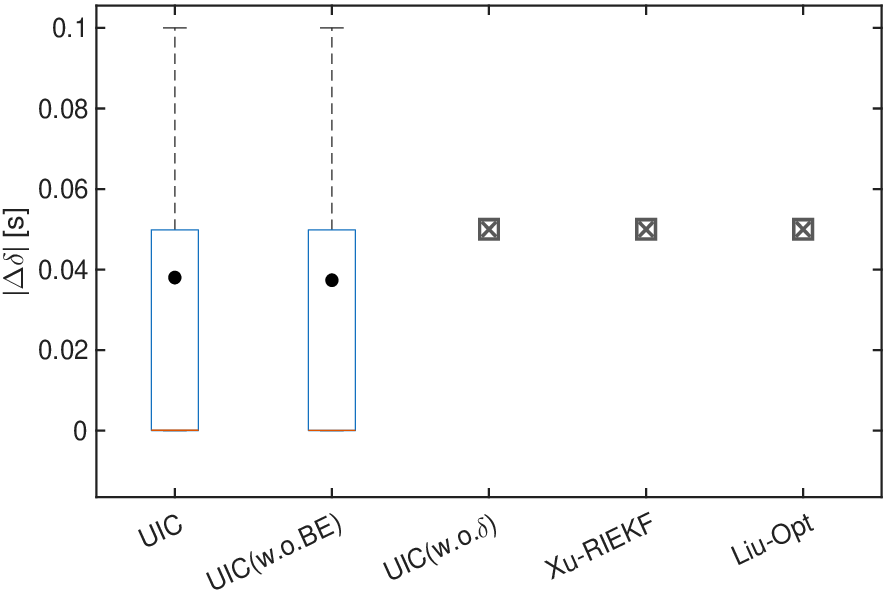}
    \label{fig:bench_tau}
  \end{subfigure}
  \hfill
  \begin{subfigure}[t]{0.245\textwidth}
    \centering
    \includegraphics[width=\linewidth]{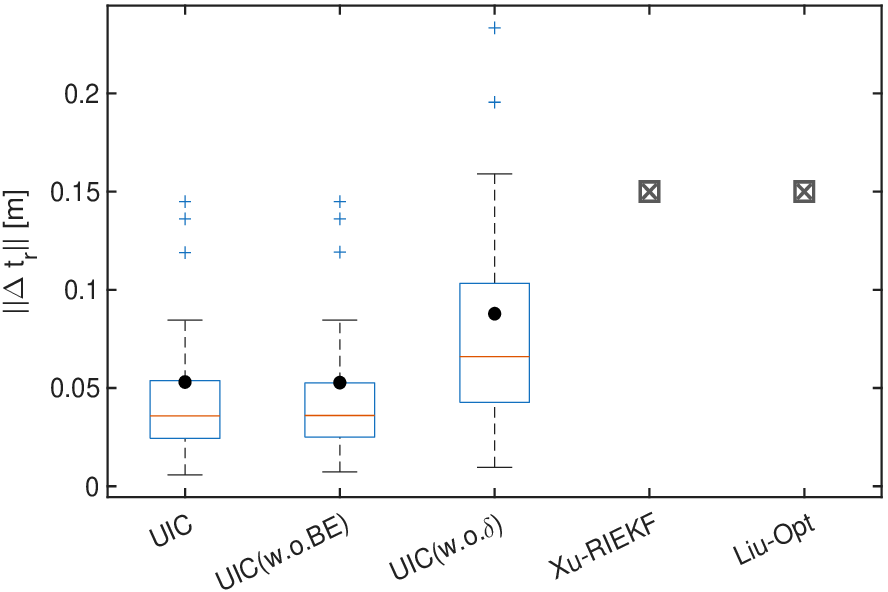}
    \label{fig:bench_tr}
  \end{subfigure}
  \hfill
  \begin{subfigure}[t]{0.245\textwidth}
    \centering
    \includegraphics[width=\linewidth]{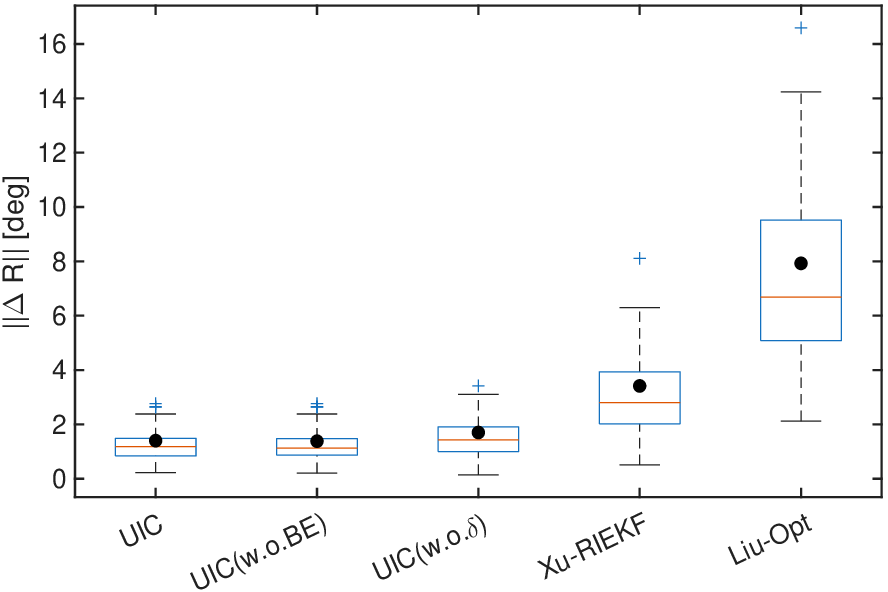}
    \label{fig:bench_rot}
  \end{subfigure}
    \hfill
\begin{subfigure}[t]{0.245\textwidth}
    \centering
    \includegraphics[width=\linewidth,height=0.66\linewidth]{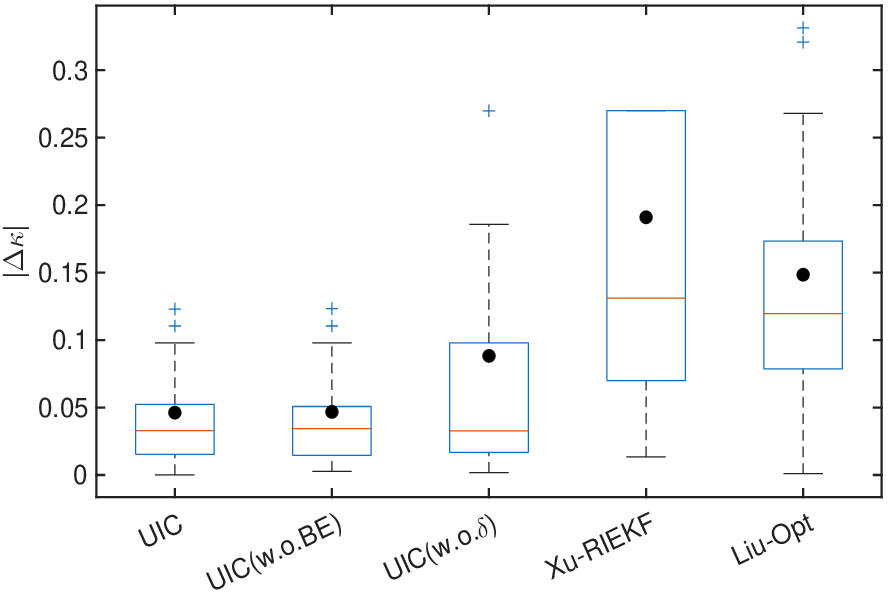}
    \label{fig:bench_kappa}
\end{subfigure}

  \caption{
Calibration Results (Absolute Error distributions of \(\Delta\delta\), \(\Delta t\), \(\Delta R\), and \(\Delta\kappa\)) and RMSE.
  Boxes show the interquartile range with the median; whiskers indicate the non-outlier range; black dots denote the RMSE.}
  \label{fig:mc_benchmark_boxplots}
\end{figure*}

\begin{figure}[t]
  \centering
  \captionsetup[sub]{justification=centering,singlelinecheck=false}

  \begin{subfigure}[t]{0.235\textwidth}
    \centering
    \includegraphics[width=\linewidth]{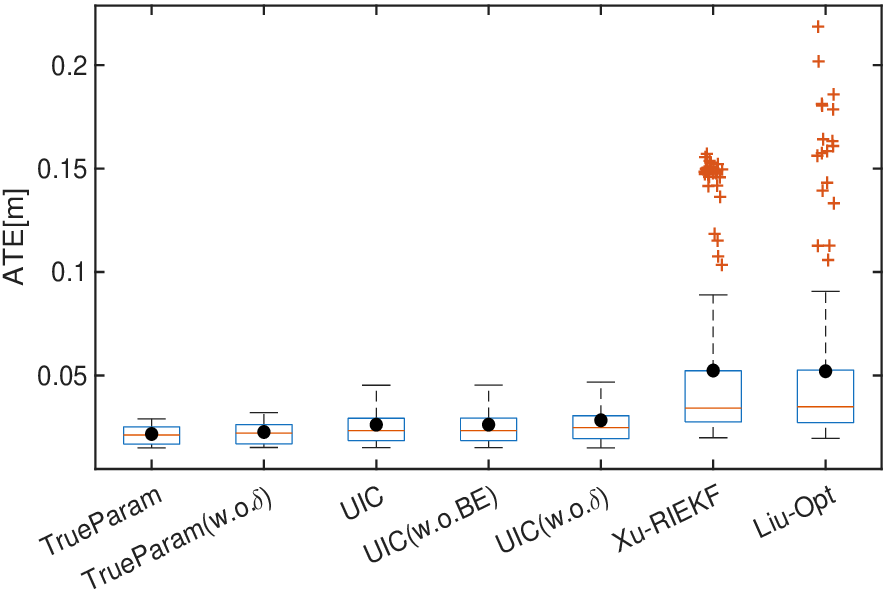}
    \label{fig:bench_tau}
  \end{subfigure}
  \begin{subfigure}[t]{0.235\textwidth}
    \centering
    \includegraphics[width=\linewidth]{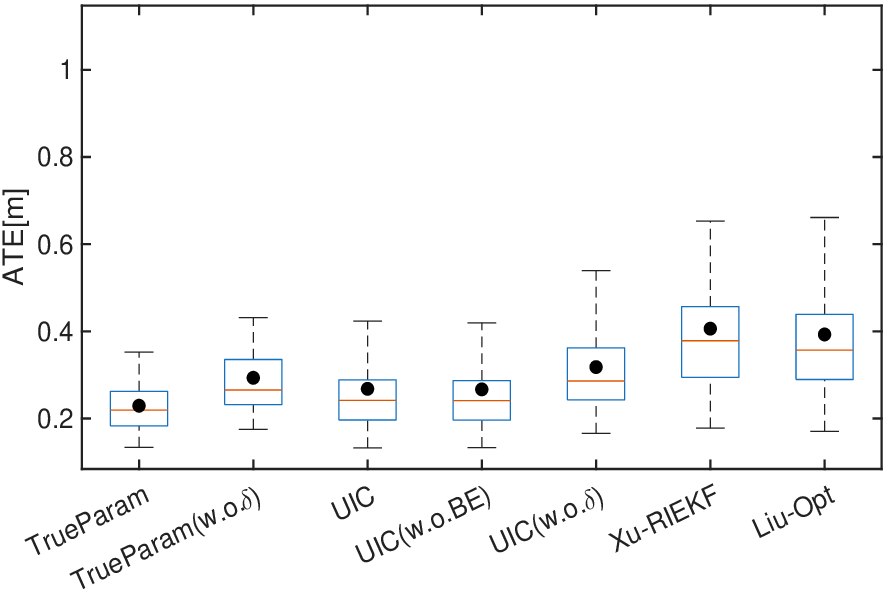}
    \label{fig:bench_tr}
  \end{subfigure}
  \caption{ATE for simple(left) and aggressive(right) test trajectories}
  \label{fig:mc_benchmark_boxplots2}
\end{figure}

\begin{figure}[t]
  \centering
  \captionsetup[sub]{justification=centering,singlelinecheck=false}
\begin{subfigure}[t]{0.5\textwidth}
    \centering
    \includegraphics[width=1\linewidth]{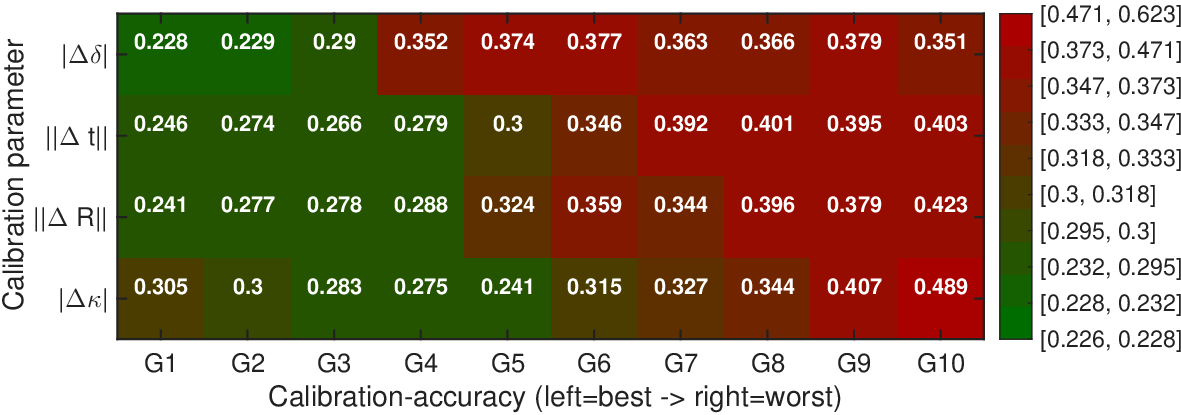}
\end{subfigure}
\begin{subfigure}[t]{0.5\textwidth}
    \centering
    \includegraphics[width=1\linewidth]{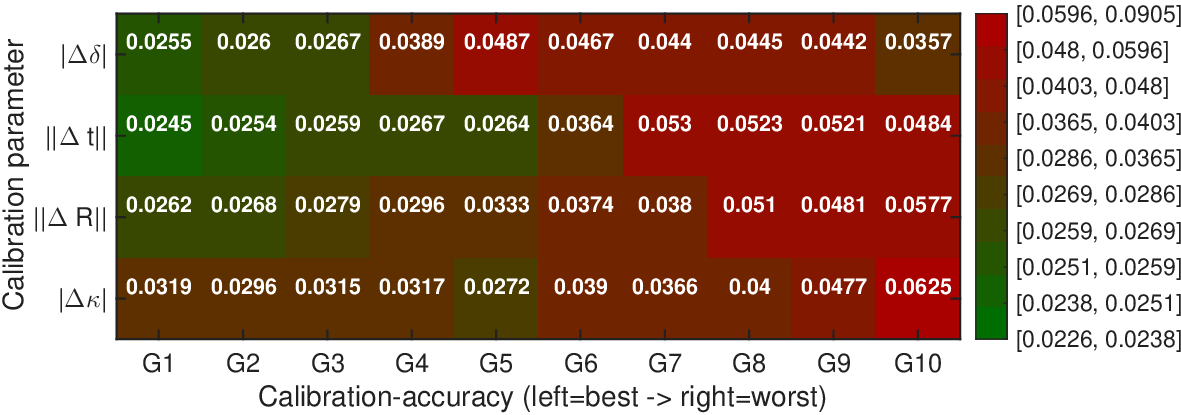}
\end{subfigure}
  \caption{Calibration accuracy vs Simple(Upper) and Aggressive(Lower) trajectory ATE(Upper)}
  \label{fig:mc_benchmark_heatmap}
\end{figure}

\subsection{UIC against robust baselines}\label{sec:mc_benchmark_baselines}
We evaluate both calibration accuracy and downstream trajectory impact in this simulation. {Calibration accuracy:} Over \(n_{\mathrm{MC}}=50\) trials, we estimate \(\hat\Theta\) and report the errors in \(\delta\)[s], \(\kappa_d\), \({}^{d}\!R_b\)[\(^{\circ}\)], and \({}^{d}\!t_b\)[m]. {Trajectory impact:} Each calibrated parameter set is evaluated on two types of independent test trajectories, each of which contains 20 different trajectories. The first type of test trajectories(Simple Trajectories) has slow motions, while the other(Aggressive Trajectories) has rapid movements. 
We run a GP-InEKF and RTS fusion that incorporates Abs/VO and DVL (with discrete-time offset compensation), and compute the ATE.

To quantify the practical gain of jointly calibrating time and space, we benchmark UIC against two reproduced robust calibration baselines for DVL, Xu-RIEKF~\cite{xu2022novel} and Liu-Opt~\cite{9911179}, which estimate only \(({}^{d}\!R_b,\kappa_d)\); thus, following their intended usage in our setting, we fix the clock offset \(\delta\) and lever arm \({}^{d}\!t_b\) to measured values when running them. We also include two UIC ablations, UIC(w.o.BE) (without bias-elimination in initialization) and UIC(w.o.\(\delta\)) (disabling temporal calibration by fixing \(\delta=0\)), under the same protocol; Figs.~\ref{fig:mc_benchmark_boxplots} and~\ref{fig:mc_benchmark_boxplots2} summarize calibration errors and the downstream ATE, respectively. In addition, to explicitly isolate the impact of temporal misalignment on trajectory estimation, we report the baseline TrueParam and TrueParam(w.o.\(\delta\)), where the spatial parameters \(({}^{d}\!R_b,{}^{d}\!t_b,\kappa_d)\) are set to their ground-truth values while \(\delta\) is set to real value and zero; therefore, any ATE degradation observed in TrueParam(w.o.\(\delta\)) can be attributed to the wrong time alignment rather than imperfect spatial calibration.

Overall, the results show that explicitly estimating a constant \(\delta\) together with \({}^{d}\!t_b\) is a key factor for robust and accurate calibration: when \(\delta\) is fixed, the remaining parameters \(({}^{d}\!R_b,\kappa_d,{}^{d}\!t_b)\) tend to absorb spatiotemporal mismatch, which degrades both calibration quality and trajectory accuracy, whereas UIC yields consistently better calibration and ATE, with the advantage being most pronounced under dynamically rich motions where temporal misalignment is more damaging; consistently, TrueParam(w.o.\(\delta\)) shows only limited ATE change on low-dynamic (Simple) trajectories but a clear degradation on high-dynamic (Aggressive) trajectories, confirming that accurate temporal calibration becomes critical when linear/angular velocities are large. The bias-elimination option provides a modest but consistent improvement for spatial parameters, but its impact is smaller than that of enabling temporal calibration and lever-arm estimation, indicating that these two components dominate this benchmark. Finally, the stratified analysis in Figs.~\ref{fig:mc_benchmark_heatmap} confirms that trajectory accuracy deteriorates as calibration quality worsens, and that the trajectory is generally more sensitive to the clock offset \(\delta\) than to the scale \(\kappa_d\).


\subsection{Angular velocity excitation ablation with fixed time offset}\label{sec:mc_angvel_tau_fixed}

This experiment isolates the effect of angular velocity excitation on the spatial DVL–IMU calibration,
with an emphasis on the lever arm \({}^{d}\!t_b\).
From section~\ref{sec:traj_design}, when angular motion is weak, the lever arm becomes weakly observable. To decouple spatial calibration from temporal misalignment, we fix the time offset to the ground truth
and disable its estimation in UIC, i.e.,
\(\delta=\delta^{o}=\SI{0.05}{s}\).
Therefore the time-offset error is identically zero (RMSE\((\Delta\delta)=0\)) by construction, and UIC estimates only
the DVL scale factor, rotation, and lever arm.

We compare three trajectory scenarios over \(n_{\mathrm{MC}}=50\) Monte Carlo runs:
(i) {HIGH\_T100}: high angular velocity excitation, \(T=\SI{100}{s}\);
(ii) {LOW\_T100}: low angular velocity excitation, \(T=\SI{100}{s}\);
(iii) {LOW\_T2000}: low angular velocity excitation, \(T=\SI{2000}{s}\).
As a scalar proxy for excitation strength, we report the mean ground-truth angular-speed magnitude
\(\overline{\lVert\omega_b\rVert}\).

\begin{table}[t]
    \centering
    \scriptsize
    \setlength{\tabcolsep}{2pt}
    \renewcommand{\arraystretch}{1.15}
    \begin{tabular}{@{}lcccccc@{}}
        \toprule
        Case &
        \(T\) [s] &
        \(\overline{\lVert\omega_b\rVert}\) [rad/s] &
        \(\Delta\kappa_d\) &
        \(\Delta R\) [deg] &
        \(\Delta t\) [m] &
        runtime [s] \\
        \midrule
        {HIGH\_T100}  & 100  & 0.602 & 0.0683 & 1.563 & 0.0976 & 0.510 \\
        {LOW\_T100}   & 100  & 0.381 & 0.0596 & 1.567 & 0.1904 & 0.507 \\
        {LOW\_T2000}  & 2000 & 0.393 & 0.0900 & 0.530 & 0.1325 & 18.35 \\
        \bottomrule
    \end{tabular}
    \caption{Summary for the angular velocity excitation ablation.}
    \label{tab:mc_angvel_excitation_tauFixed_summary}
\end{table}

Table~\ref{tab:mc_angvel_excitation_tauFixed_summary}
show that lever-arm accuracy is strongly excitation-limited.
For the same duration \(T=\SI{100}{s}\), increasing angular-rate excitation reduces the lever-arm RMSE.
In contrast, the rotation error remains around \(\approx \SI{1.56}{deg}\) for both \(\SI{100}{s}\) cases,
suggesting that in this regime \({}^{d}\!R_b\) is less sensitive to short-horizon excitation amplitude
than \({}^{d}\!t_b\). Extending a low-excitation trajectory to \(T=\SI{2000}{s}\) improves the rotation error substantially
and also improves lever-arm accuracy,
but it increases computation time markedly.

\begin{figure*}[t]
  \centering
  \captionsetup[sub]{justification=centering}
  \begin{subfigure}[t]{0.245\textwidth}
    \centering
    \includegraphics[width=\linewidth]{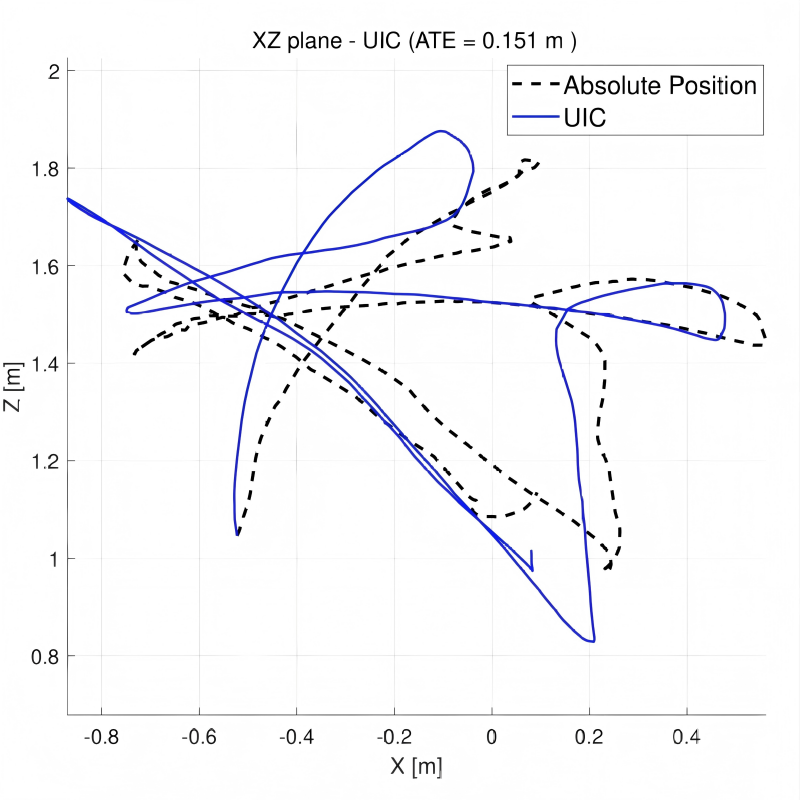}
    \caption{UIC}
  \end{subfigure}
  \begin{subfigure}[t]{0.245\textwidth}
    \centering
    \includegraphics[width=\linewidth]{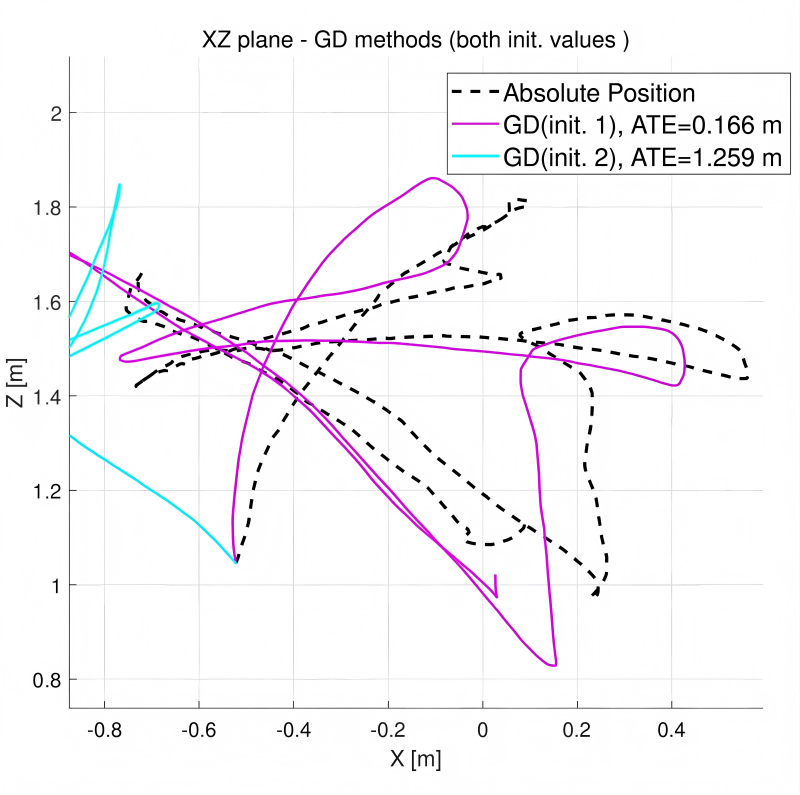}
    \caption{GD (init. 1) \& GD (init. 2)}
  \end{subfigure}
  \begin{subfigure}[t]{0.245\textwidth}
    \centering
    \includegraphics[width=\linewidth]{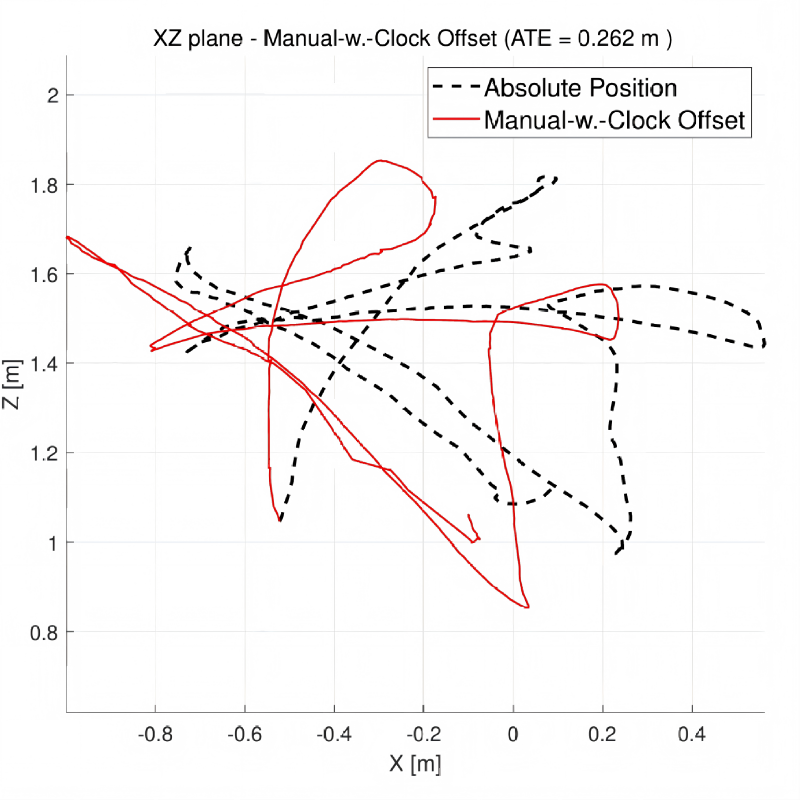}
    \caption{Manual w. $\delta_T$ }
  \end{subfigure}
  \begin{subfigure}[t]{0.245\textwidth}
    \centering
    \includegraphics[width=\linewidth]{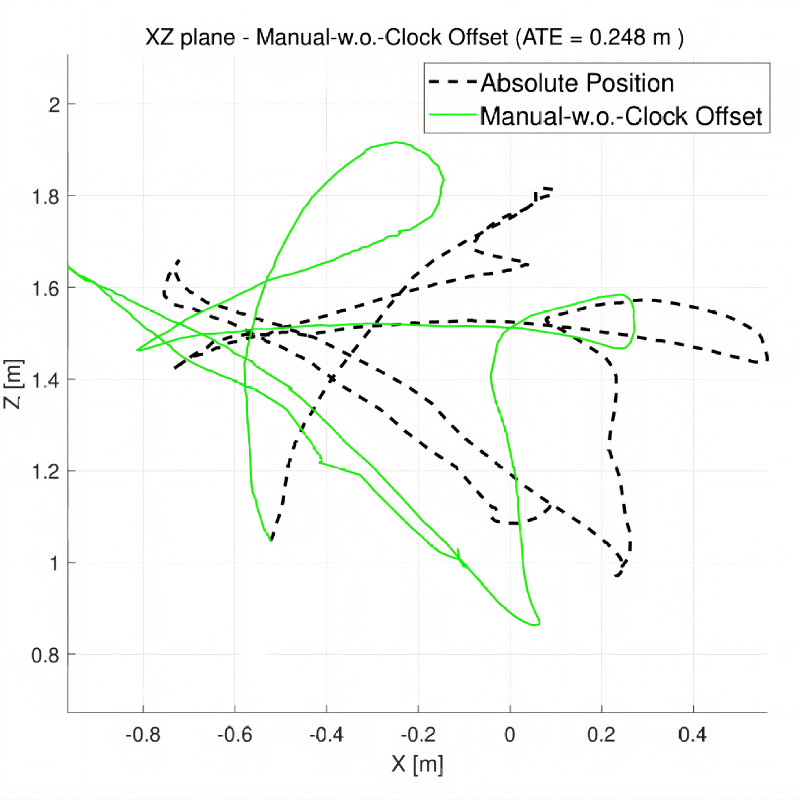}
    \caption{Manual w.o. $\delta_T$}
  \end{subfigure}
  \caption{Trajectory A estimation using the parameters calibrated from trajectory B.}
  \label{Trajectory}
\end{figure*}

\begin{figure}[t]
    \centering
    \includegraphics[width=0.9\linewidth]{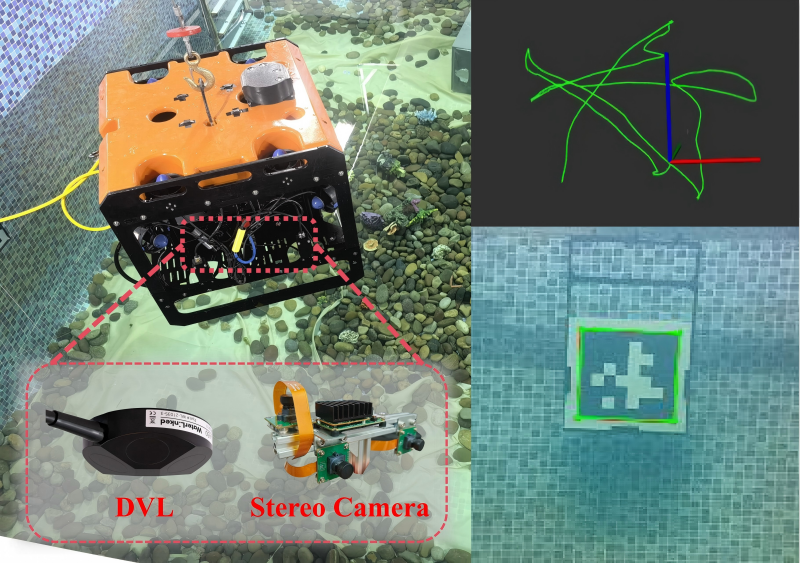}
    \caption{Water pool experiments. Left:  Underwater robot; bottom right: AprilTag board; upper right: robot trajectory.}
    \label{Experiment_setup}
\end{figure}

\begingroup
\newcommand{\best}[1]{\textbf{\textcolor{blue}{#1}}}
\newcommand{\good}[1]{\textcolor{blue!60}{#1}}

\begin{table*}[t]
\setlength{\tabcolsep}{0pt}
\renewcommand{\arraystretch}{1}
\caption{Cross-trajectory ATE and RPE. Each column pair lists the evaluation trajectory under the calibration trajectory shown above. \best{Bold blue} and \good{light blue} denote the best and second-best results. Averages are arithmetic means over the six pairs; percentages are relative to UIC and are computed from the listed per-pair values before rounding the averages.}
\label{tab:Result1}
\begin{tabular*}{1\textwidth}{@{\extracolsep{\fill}} l l *{14}{c} @{}}
\toprule
\multicolumn{2}{c}{Calibration}
  & \multicolumn{4}{c}{Traj A}
  & \multicolumn{4}{c}{Traj B}
  & \multicolumn{4}{c}{Traj C}
  & \multicolumn{2}{c}{Average}\\
\cmidrule(lr){1-2} \cmidrule(lr){3-6} \cmidrule(lr){7-10} \cmidrule(lr){11-14} \cmidrule(lr){15-16}
\multicolumn{2}{c}{Evaluate} 
  & \multicolumn{2}{c}{Traj B} & \multicolumn{2}{c}{Traj C}
  & \multicolumn{2}{c}{Traj A} & \multicolumn{2}{c}{Traj C}
  & \multicolumn{2}{c}{Traj A} & \multicolumn{2}{c}{Traj B}
  & \multicolumn{2}{c}{Comparison}\\
\cmidrule(lr){1-2} \cmidrule(lr){3-4} \cmidrule(lr){5-6}
\cmidrule(lr){7-8} \cmidrule(lr){9-10}
\cmidrule(lr){11-12} \cmidrule(lr){13-14} \cmidrule(lr){15-16}
 Method & Result
  & ATE & RPE & ATE & RPE
  & ATE & RPE & ATE & RPE
  & ATE & RPE & ATE & RPE & ATE & RPE \\
\midrule

\multirow{1}{*}{UIC} & Meter 
  & \best{0.348} & \best{0.107} & \good{0.621} & 0.156
  & \best{0.151} & \best{0.070} & \good{0.538} & \good{0.141}
  & \best{0.209} & \best{0.067} & \best{0.230} & \best{0.071}
  & \best{0.350} & \best{0.102} \\
\addlinespace[2pt]

\multirow{1}{*}{{GD}} & Meter
  & 0.368 & \best{0.107} & \best{0.606} & \good{0.154}
  & \good{0.166} & \best{0.070} & \best{0.529} & 0.142
  & \good{0.231} & \good{0.070}
  & \good{0.237} & \good{0.072}
  & \good{0.356} & \good{0.103} \\
\multirow{1}{*}{{\rm (init. 1)}}&\%
  & 105.7  & \best{100} & \best{97.58}  & \good{98.72}
  & \good{109.9}   & \best{100}   & \best{98.3}  & 100.71
  & \good{110.5}  & \good{104.5}  & \good{103.0} & \good{101.4}
  & \good{101.9}  & \good{100.5}  \\
\addlinespace[2pt]

\multirow{1}{*}{{GD}} & Meter
  & \good{0.367} & \best{0.107} & 0.993 & \best{0.144}
  & 1.259 & 0.201 & 0.942 & \best{0.138}
  & 1.158 & 0.196 & 0.750 & 0.102
  & 0.912 & 0.148 \\
\multirow{1}{*}{({init. 2})}& \%
  & \good{105.5}  & \best{100} & 159.9  & \best{92.31}
  & 833.8  & 287.1  & 175.1  & \best{97.9}
  & 554.1  & 292.5  & 326.1  & 143.7
  & 260.8  & 145.1 \\
\addlinespace[2pt]

\multirow{1}{*}{Manual} & Meter
  & 0.378 & \good{0.108} & \good{0.621} & 0.158
  & 0.262 & \good{0.085} & 0.607 & 0.157
  & 0.262 & 0.085 & 0.390 & 0.109
  & 0.420 & {0.117} \\
\multirow{1}{*}{(w. $\delta_T$)} & \%
  & 108.6  & \good{100.9}  & \good{100} & 101.3
  & 173.5  & \good{121.4}  & 112.8  & 111.3
  & 125.4  & 126.9  & 169.6  & 153.5
  & 120.2  & 114.7  \\
\addlinespace[2pt]

\multirow{1}{*}{Manual} & Meter
  & 0.388 & \good{0.108} & 0.632 & 0.158
  & 0.248 & 0.088 & 0.632 & 0.158
  & 0.248 & 0.088 & 0.388 & 0.108
  & 0.423 & 0.118 \\
\multirow{1}{*}{(w.o. $\delta_T$)}&\%
  & 111.5  & \good{100.9}  & 101.8  & 101.3
  & 164.2  & 125.7  & 117.5  & 112.1
  & 118.7  & 131.3  & 168.7  & 152.1
  & 120.9  & 115.7 \\
\bottomrule
\end{tabular*}
\end{table*}
\endgroup

\section{Real-World Experiment}
We conducted a real-world experiment to evaluate our calibration method. We recorded our calibration trajectories in a $4.5\times10\times2$m pool. An A50 DVL(around 10 Hz) is mounted on an underwater robot together with a forward-looking stereo camera(10Hz). Fig.~\ref{Experiment_setup} shows the setting of our experiment. The stereo camera provides absolute pose measurements by observing an AprilTag board affixed to the pool wall. We calibrate the camera–DVL extrinsics ${}^{d}\!R_{c}, {}^{d}\!t_{c}$, their clock offset ${}^{c}\delta_{T,d}$, and the DVL scale factor $\kappa_{d}$.

We apply GP with WNOJ\cite{Tim2019jerk} prior to estimating trajectories and manually set the kernel $Q_c$ as ${\rm diag}([100,100,100,100,100,100])$.
To evaluate the calibration performance, we construct an odometry that integrates the DVL linear velocity measurements with the stereo camera's absolute orientation measurements as follows. Let ${}^{w}\!R_{c}(T_c)$ denote the camera orientation in the world frame at time $T_c$. Over a short interval $[T_{c,1}, T_{c,2}]$ between two camera measurements, since the DVL has the same frequency as the camera, there exist two DVL measurements $\mathring{v}_{d,1}$ and $\mathring{v}_{d,2}$ at times $T_{d,1}+{}^{c}\delta_{T,d}$ and $T_{d,2}+{}^{c}\delta_{T,d}$, which satisfy $T_{d,1}+{}^{c}\delta_{T,d} \leq T_{c,1}$ and $T_{d,2}+{}^{c}\delta_{T,d} \in [T_{c,1}, T_{c,2}]$. Then, the average velocity in this interval can be approximated by $v_{d}=\dfrac{T_{d,2}+{}^{c}\delta_{T,d}-T_{c,1}}{\Delta_T}\mathring{v}_{d,1}+\dfrac{T_{c,2}-T_{d,2}-{}^{c}\delta_{T,d}}{\Delta_T}\mathring{v}_{d,2}$, where $\Delta_T=T_{c,2}-T_{c,1}$. 
Finally, the world-frame camera position update over this interval is approximated by
${}^{w}\!p_{2} \approx {}^{w}\!p_{1} + \kappa_{d}^{-1}\,{}^{w}\!R_{c}(T_{c,1})\,{}^{c}\!R_{d}\,v_{d}\,\Delta_T
+ ({ }^{w}\!R_{c}(T_{c,1}) - {}^{w}\!R_{c}(T_{c,2}))\,{}^{d}\!t_{c}$, where $\kappa_{d}^{-1}\,{}^{w}\!R_{c}(T_{c,1})\,{}^{c}\!R_{d}\,v_{d}\,\Delta_T$ and $({ }^{w}\!R_{c}(T_{c,1}) - {}^{w}\!R_{c}(T_{c,2}))\,{}^{d}\!t_{c}$ denote the updates brought by robot translation and rotation, respectively.

\begin{figure}[t]
    \centering
    \includegraphics[width=1\linewidth]{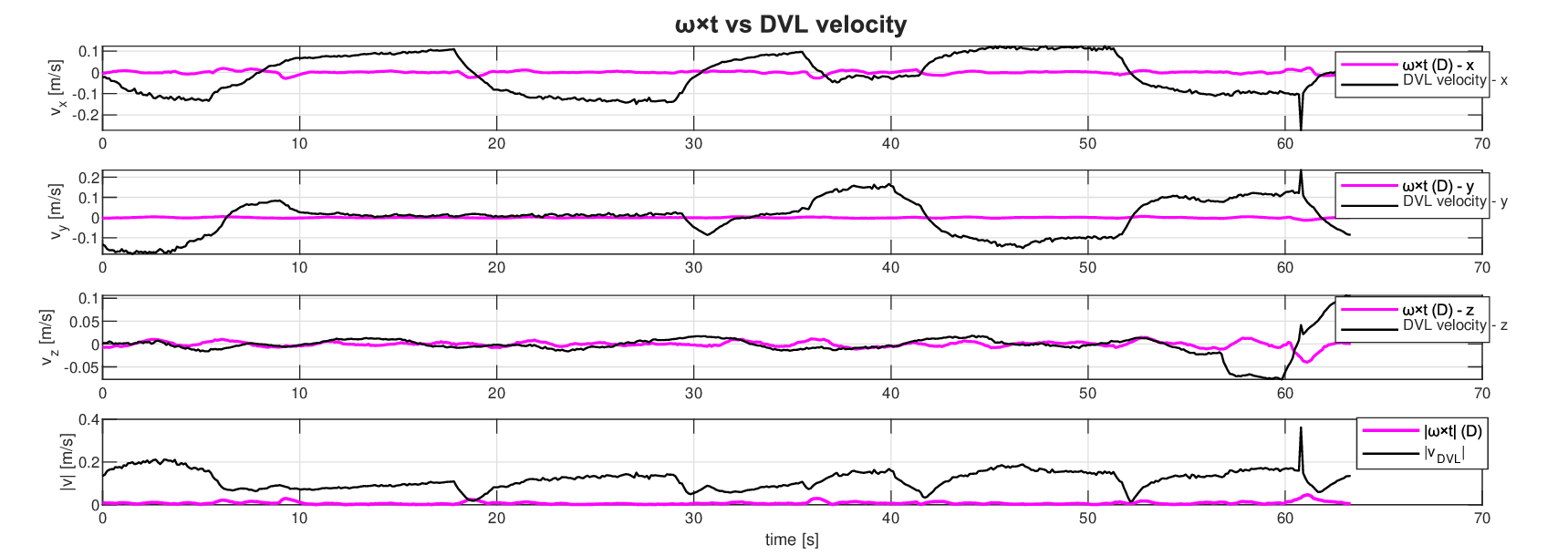}
    \caption{DVL velocity versus the lever-arm-induced term $\omega\times{}^{d}\!t_c$.}
    \label{DVL_velocity}
\end{figure}

We compare the above estimated position-only trajectory against the reference trajectory derived from the stereo camera's absolute-position measurements from the AprilTag board. Absolute Trajectory Error (ATE) and Relative Position Error (RPE) are selected as evaluation metrics.
We evaluate five methods and compare their performance:
\begin{itemize}
  \item \textbf{UIC}: our UIC method in Algorithm~\ref{DVL_algorithm}.
  \item \textbf{GD (init. 1)}: gradient decent over Problem~\eqref{problem:MAP_calibr} initialized with CAD-derived extrinsic parameters.
  \item \textbf{GD (init. 2)}: gradient decent over Problem~\eqref{problem:MAP_calibr} initialized with ${}^{d}\!R_{b}=\mathbf{I}_{3}$, ${}^{d}\!t_{b}=\mathbf{0}$.
  \item \textbf{Manual w. $\delta_T$}: use manually measured ${}^{d}\!R_{b}$ and ${}^{d}\!t_{b}$ obtained from the CAD assembly model, and compensate for the clock offset estimated by UIC.
  \item \textbf{Manual w.o. $\delta_T$}: use manually measured ${}^{d}\!R_{b}$ and ${}^{d}\!t_{b}$ obtained from the CAD assembly model, with the clock offset fixed at $0$.
\end{itemize}

We recorded three mild-motion trajectories, denoted as A, B, and C, under the same pool and camera-visibility constraints. Each trajectory is used for calibration and evaluated on the other two, yielding six cross-trajectory pairs; the results are summarized in Table~\ref{tab:Result1}, with representative trajectories shown in Fig.~\ref{Trajectory}. Trajectory C is generally more challenging for the downstream odometry: nearly all methods show larger ATE when evaluated on C, while UIC remains comparable to well-initialized GD and achieves the best average ATE and RPE without manual initialization. Poorly initialized GD is unstable, and UIC outperforms both manual baselines. The small gap between manual calibration with and without time-offset compensation is consistent with the mild motions; simulations show a larger temporal effect under aggressive dynamics.

\textbf{Limitations:} The lever arm ${}^{d}\!t_c$ is weakly observable in the pool experiment because keeping the AprilTag board within the camera's field of view limits rotational excitation; consequently, the $\omega\times{}^{d}\!t_c$ contribution in Fig.~\ref{DVL_velocity} is negligible. This is primarily an excitation limitation: richer multi-axis rotation, enabled by a wider-view or multi-target reference setup, is the preferred remedy. Priors may stabilize estimation but cannot replace missing excitation.


\section{CONCLUSION}\label{conclusion}

In this paper, we formalized DVL spatiotemporal calibration as an error-in-variables nonlinear least-squares problem under a continuous-time GP motion model, and introduced a UIC framework centered on a sequential, data-driven initialization procedure with gradient descent refinement. Under idealized conditions, we characterized the large-sample behavior of the proposed initialization estimators. Our simulations show that lever-arm estimation is excitation-limited, and the pool experiment validates the practicality of the proposed initializer under our platform constraints. Monte Carlo ablation studies further highlight that explicitly calibrating the clock offset and executing sufficiently rotationally exciting trajectories are critical for reliable spatiotemporal calibration in finite data.

\section*{Acknowledgement}
This work was supported in part by 1+1+1 CUHK-CUHK(SZ)-GDSTC Joint Collaboration Fund, JCYJ20241202124010014.
\bibliographystyle{IEEEtran}
\bibliography{references}

\end{document}